# MCE Reasoning in Recursive Causal Networks

*Wilson X. Wen*

Department of Computer Science,
The University of Melbourne,
Parkville, 3052, Australia.

## 1. Introduction

In this paper we describe a scheme for reasoning over causal networks in which known dependencies among variables can be included and multiple uncertain evidence can be present.

Our scheme is based on the principle of Minimum Cross Entropy (MCE) [5,9] and the concept of Recursive Causal Models (RCM) [4]. In this scheme:

(1) We introduce a language, the Recursive Causal Networks Description Language (RCNDL), in which known dependencies among variables can be explicitly described.

(2) We use known results about RCM to decompose the underlying probability space into subspaces, one for each RCNDL clause. Each subspace has its marginal distribution matching with the maximum likelihood estimation of the distribution of the whole space.

(3) We propagate prior information and beliefs among the clauses according to the MCE principle. For reasoning with multiple uncertain evidence, the constraint sets created by the evidence are used iteratively, and the principle of greatest gradient is used to order the constraint sets.

An RCNDL interpreter has been developed and the scheme will be incorporated into μ-Shell [14].

There has been much work on similar problems, this includes:

Lemmer's *Generalized Bayesian Updating Method* [6]. Using the Lagrange multiplier method, Lemmer derives Jeffrey's rule [3] by minimizing the cross entropy of the underlying distribution subject to one Component Marginal Distribution (CMD). This method produces an approximation of the object distribution when more than one CMD has to be considered simultaneously. When used with a tree of Local Event Groups (LEG) it can be quite efficient.

Cheeseman's *Method of Maximum Entropy* [2]. Cheeseman uses the principle of Maximum Entropy (ME) to calculate the underlying distribution given some constraints by the traditional Lagrange multiplier method. To avoid the exponential explosion of the number of states as the number of variables in the space increases, Cheeseman uses an efficient method to perform the relevant summations.

Pearl's *Method of Bayesian Networks* [8]. Pearl proposes an elegant and very efficient mechanism for propagating beliefs in parallel among the nodes in the causal networks which only needs local computation. However, it is difficult to use this method in multi-connected causal networks.

Spiegelhalter's *Method of Graphical/Recursive Models in Contingency Tables* [10]. According to the statistical theory of graphical/recursive models in contingency tables [16], this method decomposes the underlying space into subspaces and guarantees that the distributions of the subspaces

---

This research is supported by a Commonwealth Postgraduate Award and a Sigma Data Research Award in Computing.
The author's electronic address:
UUCP: wwen@mulmari.UUCP, ARPA: wwen%munnari.oz.au@uunet.uu.net,
CSNET: wwen%munnari.oz.au@australia, ACSnet: wwen@munnari.oz.au.

are the marginals of the underlying distribution. Then the beliefs for the evidence are propagated among the subspaces. Mainly Spiegelhalter discusses Bayesian evidence, and just mentions briefly that uncertain evidence is handled by introducing some extra nodes. It is not clear how this method would deal with the case of multiple uncertain evidence [11]. In his method, Spiegelhalter also uses an updating rule which is very similar to Jeffrey's Rule.

## 2. The Principle of Minimum Cross Entropy

Suppose that a system $S$ of $m$ binary discrete random variables $x_i$ ($i$=0,...,$m-1$) has a set of $2^m$ possible states $\{s_j \mid 0 \le j < 2^m\}$ with unknown (underlying) distribution $p=\{P(s_j)\}$, and we know some constraints and a prior distribution $p^{(0)}$ that estimates $p$. According to the MCE principle [5,9], the best estimate $\hat{p}$ of $p$ that satisfies the constraints is the one with the least cross entropy

$$CE(\{s_j\}) = CE(\hat{p}, p^{(0)}) = \sum_{j=0}^{2^m-1} \hat{P}(s_j) \log \frac{\hat{P}(s_j)}{P^{(0)}(s_j)}. \qquad (2.1)$$

Here we mainly consider marginal, conditional, and linear equality constraint problems, and present algorithms for some other problems such as moment constraint problems elsewhere [12].

### 2.1. Marginal Constraints and Jeffrey's Rule

According to the values of $n$ ($n<m$) distinct variables, $x_{i_k}$ ($0 \le i_k < m$, $k$=0,...,$n-1$) in $S$, we partition the state space $\{s_j\}$ into $2^n$ exclusive and exhaustive subspaces called *events* $S_l$, $l$=0,...,$2^n-1$ such that in each of these events the value of the vector $<x_{i_0}, ..., x_{i_{n-1}}>$ is fixed.

Suppose we know the *marginal constraints*, i.e. probabilities $\{P(S_l) \mid l=0,...,2^n-1\}$, for all these events from some evidence. It can be shown by the traditional Lagrange multiplier method ( see, for example, [6, 12] ) that the MCE posterior distribution which satisfies the constraint set $\{P(S_l)\}$ is

$$\hat{P}(s_j) = P^{(0)}(s_j) \frac{P(S_l)}{P^{(0)}(S_l)} \qquad (\textit{for } l=0,...2^n-1 \textit{ and } j=0,...,2^m-1, \textit{ if } s_j \in S_l) \qquad (2.2)$$

This is equivalent to Jeffrey's rule [3] $\hat{P}(s_j) = P^{(0)}(s_j \mid S_l)\, P(S_l)$, where $s_j \in S_l$.

If all these constraints are 0 or 1, the corresponding evidence is called *Bayesian evidence* (the constraint set is a Bayesian constraint set), otherwise *uncertain evidence* (an uncertain constraint set). If two or more uncertain constraint sets are simultaneously created by the evidence, the corresponding reasoning problem is called reasoning with *multiple uncertain evidence*.

### 2.2. Conditional Constraint Problems

Following the definitions in the previous section, we partition each $S_l$ further into two exclusive and exhaustive events $S_{l_0}$ and $S_{l_1}$ according to another variable $x_{i_n}$ in $S$, ($x_{i_n} \notin \{x_{i_k}\}$), such that the values of $x_{i_n}$ are 0 and 1 in $S_{l_0}$ and $S_{l_1}$, respectively.

Suppose in addition to $p^{(0)}$ we also know the *conditional constraints* $\{P(x_{i_n} \mid S_l)\}$. It can be shown ( [12], see also [17] for some more general cases) that

$$\hat{P}(s_j) = P^{(0)}(s_j) \left( \frac{P(\neg x_{i_n} \mid S_l)\, P^{(0)}(S_{l_1})}{P(x_{i_n} \mid S_l)\, P^{(0)}(S_{l_0})} \right)^{\alpha}, \textit{ where } \alpha = \begin{cases} P(x_{i_n} \mid S_l), & s_j \in S_{l_0}, \\ P(\neg x_{i_n} \mid S_l), & s_j \in S_{l_1}. \end{cases} \qquad (2.3)$$

### 2.3. Linear Constraint Problems and Numerical Techniques

A Linear Equality Constraint (LEC) problem

$$\left\{ \begin{array}{lll} \textit{Minimize} & (2.1), & \\ \textit{subject to} & \sum_{j=0}^{2^m-1} a_{kj}\, \hat{P}(s_j) = b_k, & k = 0,\ldots,n-1. \end{array} \right.$$

is equivalent to minimization of the following *Dual function* (see [7, 12])

$$D(\lambda) = \sum_{j=0}^{2^m-1} P^{(0)}(s_j)\, e^{\beta} + \sum_{k=0}^{n-1} \lambda_k\, b_k, \qquad \textit{where } \beta = -(\sum_{k=0}^{n-1} \lambda_k\, a_{kj} + 1).$$

and $\lambda=(\lambda_0, \ldots, \lambda_{n-1})$ is the vector of Lagrange multipliers. The gradients of the Dual are

$$\nabla D_k = D'(\lambda_k) = -\sum_{j=0}^{2^m-1} a_{kj}\, P^{(0)}(s_j)\, e^{\beta} + b_k = -\sum_{j=0}^{2^m-1} a_{kj}\, \hat{P}(s_j) + b_k. \tag{2.4}$$

In marginal constraint cases, these are merely the differences between the values of the constraints and the corresponding prior (see examples in section 6). Having obtained the gradients, the problem of minimization can be easily solved by Fletcher-Reeves method [7].

## 3. Recursive Causal Networks

According to Kiiveri et al [4] *Recursive Causal Model* (RCM) is characterized by an ordering of the variables $x_0,\ldots,x_{m-1}$ and the following factorization of the joint distribution:

$$P(x_0,\ldots,x_{m-1}) = P(x_0,\ldots,x_{r-1}) \prod_{r\le j<m} P(x_j \mid D_j), \tag{3.1}$$

where $D_j=\{x_{j_0},\ldots,x_{j_k}\}$, $0\le j_0<\ldots<j_k<j$. The set $S_{root} = \{x_0,\ldots,x_{r-1}\}$ is called the *root* of the model.

A *Recursive Causal Network* (RCNet) is a directed acyclic graph $<V,E>$ where the nodes in $V$ represent the variables of an RCM and the links from the nodes in $D_j$ to the node $x_j$ ($D_j \subset V$, $x_j \in V$) represent the dependencies (conditional probabilities $P(x_j \mid D_j)$) of $x_j$ on the nodes in $D_j$.

For the well known example of Cooper:

> Metastatic cancer (A) is a possible cause of a brain tumor (B) and is also an explanation for increased total serum calcium (C). In turn, either of these could explain a patient falling into a coma (D). Severe headache (E) is also possibly associated with a brain tumor.

We have the following RCM with $S_{root} = \{A\}$

$$P(A,B,C,D,E) = P(A)\,P(B \mid A)\,P(C \mid A)\,P(D \mid B,C)\,P(E \mid C).$$

where the initial description of the distribution is

$P(A) = 0.2$

| | |
|---|---|
| $P(B \mid A) = 0.8$ | $P(B \mid \neg A) = 0.2$ |
| $P(C \mid A) = 0.2$ | $P(C \mid \neg B) = 0.05$ |
| $P(D \mid B,C) = 0.8$ | $P(D \mid B,\neg C) = 0.8$ |
| $P(D \mid \neg B,C) = 0.8$ | $P(D \mid \neg B,\neg C) = 0.05$ |
| $P(E \mid C) = 0.8$ | $P(E \mid \neg C) = 0.6$ |

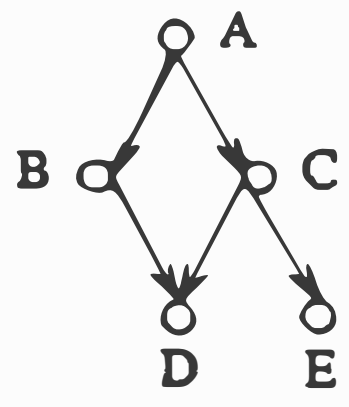

Fig. 3.1

The corresponding RCNet is shown in Fig. 3.1.

From (3.1), the cross entropy of the joint probability can be expressed as [15, 17]:

$$CE(S) = CE(S_{root}) + \sum_{r\le j<m} (CE(\bar{D}_j) - CE(D_j)) \tag{3.2}$$

where $\bar{D}_j = D_j \cup \{x_j\}$. This corresponds to Lemmer's tree of LEG's (see [6] Fig. 1). In (3.2), $S_{root}$ can be decomposed further into its cliques $C_x$. This singly connected network of LEG's corresponds to an RCNet and has the following properties:

(1) **Parallel Intersection** [17]: If we know all $P(D_j)$'s then the minimum $CE(S)$ is obtained by minimizing $CE(S_{root})$ and each $CE(\bar{D}_j)$ separately subject to $P(D_j)$ and any other constraints.

(2) **Running Intersection:** If there are constraints in only one of the sets $S_{root}$ and $\bar{D}_j$'s, then the minimum $CE(S)$ is obtained by minimizing one of $CE(S_{root})$ and $CE(\bar{D}_j)$'s subject to the constraints and, step by step, minimizing the others subject to the $P(D_j)$'s obtained.

These properties are very important. Because of them we may decompose the whole space into subspaces and propagate the beliefs among the subspaces in order to avoid the exponential explosion of the number of states in the space.

## 4. Reasoning with Multiple Uncertain Evidence

According to Brown [1], (2.2) can be also used as an approximation for reasoning with multiple uncertain evidence if we use the constraint sets one at a time. Brown proved that

(P1) The approximation at each step is a unit sum distribution, i.e. it is consistent by itself.

(P2) The approximation improves at each step according to the MCE principle. That is, the value of cross entropy decreases at each step.

(P3) The procedure converges to the MCE solution of the reasoning with multiple uncertain evidence.

Equation (2.3), when used as an approximation for a problem of multiple uncertain evidence, has properties similar to P2 and P3 above. Property P1 is obtained by multiplying a normalization factor [12]. As pointed out by Brown, after one constraint set has been used to update the approximation, the constraint sets used before may no longer be satisfied, and the last constraint set dominates. So some constraint sets may need to be used more than once to attain convergence. The constraint sets need not be used in any particular order, but the order will have an effect on the rate of convergence. Similar to strategies used in conventional nonlinear optimization [7], we have found that if we use the constraint sets in the order of *greatest gradient*, we can significantly speed up convergence.

Our scheme uses the following gradient-threshold method to control the termination of the iteration and the precision of the result:

(1) For those constraint sets that we really don't want to be washed out by other constraint sets (see Appendix C of [17] for an example), give them zero thresholds or small ones.

(2) For those which are not very important, specify large or even unit thresholds.

(3) At each iterative step, the inference system checks and updates the gradients of the constraint sets and uses the constraint set with the greatest gradient.

(4) When the gradients of all the constraint sets are smaller than the thresholds specified beforehand, the iteration terminates.

This method is very similar to the Gauss-Southwell method [7], thus can be expected to converge linearly and with a ratio close to that of the steepest descent method (see [12]).

## 5. A Description Language for Recursive Causal Networks: RCNDL

In this section, we present a description language, RCNDL, for RCNets. An RCNDL program consists of a set of clauses:

*program* ::= {*clause*} (5.1)

There are three types of clauses corresponding to the three types of structures in (3.1). These are

*clause* ::= *?–query.* | (5.2)
*head→body.* | (5.3)
*observations.* (5.4)

(1) **Queries** correspond to the root of the RCNet or $S_{root}$ in section 3. In (5.2), *query* is a list of pairs each of which consists of a set of propositions corresponding to a clique in $S_{root}$ and a list of probabilities -- the joint prior of the clique, which should be known before the query.

*query* ::= *{proposition_list : prior;}proposition_list : prior*
*proposition_list* ::= *{proposition ,} proposition*
*proposition* ::= *identifier*
*prior* ::= *pr_list*
*pr_list* ::= [ *{pr ,} pr* ]
*pr* ::= *expression.*

In most cases, the expressions are simply real numbers in [0.0, 1.0]. In the case of incomplete information, -1.0 is allowed in pr_list to represent unknown probability.

(2) Each inference rule (5.3) corresponds to a set of links from the nodes in $D_j$ (*head*) to the node $x_j$ (*body*) in the RCNet (the conditional probabilities $P(x_j \mid D_j)$ in (3.1)).

*head* ::= *proposition_list*
*body* ::= *proposition_list : pr_list*

and the list *pr_list* contains the conditional probabilities corresponding to $P(x_j \mid D_j)$ in (3.1).

(3) Observations correspond to the leaves or terminal nodes in the RCNet. Here, we have

*observations* ::= *proposition_list*

where *propositions* are the variables to be observed which form a constraint set.

An RCNDL interpreter in Prolog has been developed [15]. It has two phases:

(1) In the first phase, a preprocessor converts the RCNDL source into an intermediate form. According to (2.2) and (2.3) or section 2.3, the prior information is propagated from $S_{root}$ around the network to give a complete description to the network. After preprocessing, each clause in the intermediate program is followed by a list of the joint prior probabilities of all variables in its head and body.

(2) The reasoning under uncertainty is accomplished in the second phase of the interpreter. The constraint sets (eg. marginal, conditional, expectation, or even moment constraints) on the observed variables are ordered and propagated from clause to clause if their gradients (see (2.4)) are greater than the corresponding thresholds. If all the gradients are less than the corresponding thresholds the reasoning phase stops and the result is reported.

## 6. Examples

In this section, two simple examples are given to show how reasoning under uncertainty with the RCNDL interpreter is accomplished. For more complicated examples, see [13, 15].

### 6.1. A Simple Example of Reasoning with Multiple Uncertain Evidence

Suppose we have a simple recursive causal model $\{A, B, C\}$ (Fig. 6.1) as follows

$P(A, B, C) = P(A)\,P(B \mid A)\,P(C \mid A)$, *and*

$P(A) = 0.700000$,

$P(B \mid \neg A) = 0.200000$, $P(B \mid A) = 0.400000$,

$P(C \mid \neg A) = 0.800000$, $P(C \mid A) = 0.100000$.

We have the following RCNDL program

$? - A$ : [0.300000, 0.700000].
$A \rightarrow B$ : [0.200000, 0.400000].
$A \rightarrow C$ : [0.800000, 0.100000].
$B$.

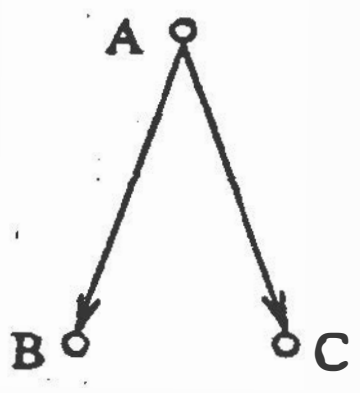

Fig. 6.1

$C$.

After preprocessing [15], we have the following intermediate form of the program

$A \to B$ : [0.240000, 0.060000, 0.420000, 0.280000].

$A \to C$ : [0.060000, 0.240000, 0.630000, 0.070000].

$B$ : [0.660000, 0.340000].

$C$ : [0.690000, 0.310000].

Suppose we know constraints (for first order marginal distributions, we only talk about constraints instead of constraint sets) $P(B) = 0.330000$ and $P(C) = 0.950000$, Because $|\nabla C| = 0.640000 > |\nabla B| = 0.010000$, we choose $P(C)$ to update the distribution $\{A, C\}$ first. By (2.2), we obtain a new clause

$A \to C$ : [0.004348, 0.735484, 0.045652, 0.214516].

and $P(A) = 0.260168$. Then using it to update $\{A, B\}$ we get

$A \to B$ : [0.591866, 0.147966, 0.156101, 0.104067].

where, $P(B) = 0.252034$, and $|\nabla B| = 0.077966$. If we have specified a threshold 0.01 beforehand, then we have to use the remaining constraint P(B) to update the result again and obtain

$A \to B$ : [0.530171,0.193740,0.139829,0.136260],

which gives $P(A) = 0.276089$. Continue updating $\{A, C\}$ by P(A), we have

$A \to C$ : [0.004254,0.719657,0.048446,0.227643]

where $P(C) = 0.947300$. $|\nabla(C)| = 0.002700$ is less than 0.01 which means we may stop here. Comparing the result $P(A) = 0.276089$ here with the MCE result of $P(A) = 0.274364$, the error is 0.001725. If we have a smaller threshold, say 0.001, then we need one more pass (by one pass, we mean using the constraints or constraint sets, once each), so that a more accurate result $P(A) = 0.274341$ is obtained. The error here is 0.000023 and $|\nabla(C)| = 0.000014$.

If we use constraint $P(B)$ first, even for the threshold 0.01, two passes will be needed to obtain a result of $P(A) = 0.274248$ with an error of 0.000116 and $|\nabla(B)| = 0.000461$.

For the same prior distribution, several cases of different constraints are given in Table 6.1:

Table 6.1 More Constraint Problems for Example 6.1

| Constraints | | | | Ps(A) | | | |
|---|---|---|---|---|---|---|---|
| P(B) | gradient | P(C) | gradient | step more | P(B) first | P(C) first | MCE value |
| 0.330000 | -0.010000 | 0.950000 | 0.640000 | 1.<br>2. | 0.290038<br>0.274248 | 0.276089<br>0.274341 | 0.274364 |
| 1.000000 | 0.660000 | 0.150000 | -0.160000 | 1.<br>2. | 0.866627<br>0.866627 | 0.895002<br>0.866627 | 0.866537 |
| 0.150000 | -0.160000 | 0.670000 | 0.360000 | 1.<br>2. | 0.429631<br>0.435663 | 0.418813<br>0.433291 | 0.433053 |
| 0.270000 | -0.070000 | 0.050000 | -0.260000 | 1.<br>2. | 0.873383<br>0.870505 | 0.869070<br>0.871116 | 0.871064 |
| 0.650000 | 0.310000 | 0.850000 | 0.540000 | 1.<br>2. | 0.379245<br>0.398768 | 0.415431<br>0.393405 | 0.394492 |
| 0.950000 | 0.610000 | 0.850000 | 0.540000 | 1.<br>2. | 0.443543<br>0.448283 | 0.457625<br>- | 0.447418 |

It is easy to see from Table 6.1 that for this example the accuracy of our method is quite satisfactory and the greatest gradient principle for selection of constraints improves the accuracy.

### 6.2. The Example of Metastatic Cancer

For Cooper's example (Fig. 3.1), We may write the corresponding RCNDL program as follows

$?-A$ : [0.800000, 0.200000].

$A \rightarrow B$ : [0.200000, 0.800000].

$A \rightarrow C$ : [0.050000, 0.200000].

$B, C \rightarrow D$ : [0.050000, 0.800000, 0.800000, 0.800000].

$C \rightarrow E$ : [0.600000, 0.800000].

$D$.

$E$.

If the observed probabilities of D and E are 0 and 1, respectively, only one pass is needed to obtain the correct result $P(A) = 0.097278$. Both gradients of the Bayesian constraints $P(D)$ and $P(E)$ become zero after one pass of updating.

If we have uncertain constraints $P(D) = 0.750000$, $P(E) = 0.100000$ and we use the constraint $P(E)$ first due to the greater gradient, we get the result $P(A) = 0.336083$, which is very close to the MCE result 0.336007, in one pass. However, if we use constraint $P(D)$ first, two passes will be needed to obtain a result with similar accuracy.

## 7. Conclusions

In this paper, relationships between the MCE principle and the RCM concept are investigated. An RCNet can be decomposed into small pieces and the joint distribution of the RCNet matches with the marginal distributions of the pieces perfectly in the sense of MCE reasoning with a single constraint set or Bayesian constraint sets. The problem of multiple uncertain evidence is solved by using the constraint sets one at a time iteratively, and the convergence can be speeded up by careful ordering of the constraint sets. An overall scheme of MCE reasoning in RCNet is proposed based on the above analysis. The dependency and correlations among the variables are described in a special language RCNDL. An interpreter for RCNDL language has been developed. The performance of the scheme is illustrated on two well known examples.

Our method overcomes the computational difficulty in probabilistic reasoning in the case of large sparse probability space. The number of states for which the probabilities are evaluated in our method is less than $2^n k$, where $n$ is the number of variables in the largest clauses in the RCNDL program and $k$ is the number of clauses in the program. For the conventional MCE method the number of states to be evaluated is equal to $2^m$.

The efficiency of the method is quite satisfactory because of the linear convergence of the steepest gradient method. In our experiences, in most cases the method produces quite accurate results in a few passes of updating. For Bayesian evidence it needs only one pass to produce the right result [12, 17]. In this case, it has Spiegelhalter's method as a special case, for which it is not clear how to handle the case of multiple uncertain evidence.

Our method is similar to Lemmer's method [6], which handles trees of LEG's, but ours also handles singly connected RCNets [13]. Lemmer's method for selecting CMD's seems not as parsimonious as ours. Spiegelhalter [10] uses an efficient "filling out" algorithm, to convert the graph of the underlying space into a triangulated graph which is actually equivalent to the concept of decomposable models in statistics. This method seems even more parsimonious than ours, but with penalty of loss of explicit causality in the original graph.

It is not difficult to generalize our method to include some small directed cycles in one RCNDL clause because such cycles are not excluded by our principle [15, 17] which actually only needs independence among the clauses. It is possible to modify our method to implement a parallel MCE reasoning mechanism for singly connected networks of LEG's . Actually, our method is more suitable for parallel computation and a parallel RCNDL interpreter [13] has been developed for the Encore computer system, a shared memory multiprocessor system.

This paper is a simplified version of [15]. The reader who is interested in other types of constraint problems should refer [12] for detail.

**Acknowledgement** Thanks to E. A. Sonenberg, B. Marksjo, R. Watson, G. Port, and P. Cheeseman for valuable discussions, instructive comments, encouragement and helpfulness. Thanks also to R. Quinlan, J. Pearl, J. Lemmer and B. Wise for instructive advice and comments.